\documentclass[letterpaper, 10 pt, conference]{ieeeconf}
\IEEEoverridecommandlockouts  
\usepackage{cite}

\usepackage{amsmath,amssymb,amsfonts}
\usepackage{algorithmic}
\usepackage{graphicx}
\usepackage{textcomp}
\usepackage{xcolor}
\usepackage{amsmath}
\usepackage{amssymb}
\usepackage{mathtools}

\usepackage{amsthm}
\usepackage{algorithm}
\usepackage{algorithmic}
\usepackage{hyperref}
\usepackage[utf8]{inputenc} % allow utf-8 input
\usepackage[T1]{fontenc}    % use 8-bit T1 fonts、
\usepackage{hyperref}       % hyperlinks
\usepackage{url}            % simple URL typesetting
\usepackage{booktabs}       % professional-quality tables
\usepackage{amsfonts}       % blackboard math symbols
\usepackage{nicefrac}       % compact symbols for 1/2, etc.
\usepackage{microtype}      % microtypography
\usepackage{xcolor}         % colors
\usepackage{microtype}
\usepackage{graphicx}
\usepackage{subfigure}
\usepackage{makecell}
\usepackage{authblk}
\usepackage{multirow}
\usepackage{rotating}
\usepackage{booktabs}
\usepackage{threeparttable}
\usepackage{algorithm}
\usepackage{algorithmic}
\usepackage{cite}
\usepackage{amsmath,amssymb,amsfonts}
\usepackage{algorithmic}
\usepackage{graphicx}
\usepackage{textcomp}
\usepackage{xcolor}
\usepackage{amsmath}
\usepackage{amssymb}
\usepackage{mathtools}
\usepackage{multirow}

\usepackage{amsthm}
\usepackage{algorithm}
\usepackage{algorithmic}
\usepackage{hyperref}
\usepackage[utf8]{inputenc} % allow utf-8 input
\usepackage[T1]{fontenc}    % use 8-bit T1 fonts、
\usepackage{hyperref}       % hyperlinks
\usepackage{url}            % simple URL typesetting
\usepackage{booktabs}       % professional-quality tables
\usepackage{amsfonts}       % blackboard math symbols
\usepackage{nicefrac}       % compact symbols for 1/2, etc.
\usepackage{microtype}      % microtypography
\usepackage{xcolor}         % colors
\usepackage{microtype}
\usepackage{graphicx}
\usepackage{subfigure}
\usepackage{makecell}

     \PassOptionsToPackage{numbers, compress}{natbib}
\def\BibTeX{{\rm B\kern-.05em{\sc i\kern-.025em b}\kern-.08em
    T\kern-.1667em\lower.7ex\hbox{E}\kern-.125emX}}
\begin{document}

% to compile a preprint version, e.g., for submission to arXiv, add add the
% [preprint] option:
%     \usepackage[preprint]{neurips_2023}

% to compile a camera-ready version, add the [final] option, e.g.:
%     \usepackage[final]{neurips_2023}

% to avoid loading the natbib package, add option nonatbib:
%    \usepackage[nonatbib]{neurips_2023}

%\title{Learning Dual-arm Object Rearrangement for Cartesian Robots with Reinforcement Learning}
\title{Learning Flexible Job Shop Scheduling under Limited Buffers and \\ Material Kitting Constraints}
% \author{Shishun Zhang$^{1}$, Qijin She$^{1}$, Wenhao Li$^{1}$, Chenyang Zhu$^{1}$, Yongjun Wang$^{1}$, Ruizhen Hu$^{3}$, Kai Xu$^{1,2,*}$ \\
% $^{1}$National University of Defense Technology $\quad$ $^{2}$Xiangjiang Laboratory $\quad$ $^{3}$Shenzhen University \\
% $^{*}$Corresponding Author}
%\author{\authorblockA}
%\author{Shishun Zhang$^{1}$, Qijin She$^{1}$, Wenhao Li$^{1}$, Chenyang Zhu$^{1}$, Yongjun Wang$^{1}$, Ruizhen Hu$^{2}$, Kai Xu$^{*1,3}$% <-this % stops a space
%\thanks{$^{*}$corresponding author.}
%\thanks{This work was supported in part by the NSFC (62325211, 62132021), the National Key Research and Development Program of China (2018AAA0102200) and the Major Program of Xiangjiang Laboratory (23XJ01009).}% <-this % stops a space
%\thanks{$^{1}$Shishun Zhang, Qijin She, Wenhao Li, Chenyang Zhu, Yongjun Wang and Kai Xu are with the Department of Computer Science, National University of Defense Technology.}%
%\thanks{$^{2}$ RuizhenHu is with the Department of Computer Science, Shenzhen University.}%
%}
\author{Shishun Zhang$^{1}$, Juzhan Xu$^{3}$, Yidan Fan$^{1}$, Chenyang Zhu$^{1}$, Ruizhen Hu$^{3}$, Yongjun Wang$^{1}$, Kai Xu$^{2,*}$ \\
$^{1}$National University of Defense Technology $\quad$ $^{2}$Institute of AI for Industries, Chinese Academy of Sciences \\ $^{3}$Shenzhen University $\quad$
$^{*}$Corresponding Author}

\maketitle

\begin{abstract}
	The Flexible Job Shop Scheduling Problem (FJSP) originates from real production lines, while some practical constraints are often ignored or idealized in current FJSP studies, among which the limited buffer problem has a particular impact on production efficiency. To this end, we study an extended problem that is closer to practical scenarios—the Flexible Job Shop Scheduling Problem with Limited Buffers and Material Kitting. In recent years, deep reinforcement learning (DRL) has demonstrated considerable potential in scheduling tasks. However, its capacity for state modeling remains limited when handling complex dependencies and long-term constraints. To address this, we leverage a heterogeneous graph network within the DRL framework to model the global state. By constructing efficient message passing among machines, operations, and buffers, the network focuses on avoiding decisions that may cause frequent pallet changes during long-sequence scheduling, thereby helping improve buffer utilization and overall decision quality. Experimental results on both synthetic and real production line datasets show that the proposed method outperforms traditional heuristics and advanced DRL methods in terms of makespan and pallet changes, and also achieves a good balance between solution quality and computational cost. Furthermore, a supplementary video is provided to showcase a simulation system that effectively visualizes the progression of the production line.
\end{abstract}

\section{Introduction}

The Flexible Job-Shop Scheduling Problem (FJSP) is a core optimization challenge in modern manufacturing and has received increasing research attention in recent years \cite{gao2019review,jqr-47-4-581}. The main challenge lies in simultaneously optimizing the operation sequence and machine allocation decisions. As the number of machines and operations grows, the solution space becomes vast, making it a strong NP-hard problem. Recently, with the continuous development of Operations Research (OR) and Machine Learning (ML) fields, FJSP has gradually been addressed with improved solution methods. %%can reduce fjsp illustration

However, the standard FJSP often overlooks complex resource constraints prevalent in real-world production lines. In high-mix scenarios such as steel plate processing and part sorting, parts must be temporarily stored in a limited number of buffer zones (e.g., pallets) under strict material kitting rules, where each pallet can only accommodate parts of the same category \cite{zhang2022material}. This results in a part-sorting bottleneck: insufficient pallets for diverse part types lead to frequent pallet changes, causing congestion and efficiency losses. This gives rise to the problem, we define the Flexible Job-Shop Scheduling Problem with Limited Buffers and Material Kitting (\textbf{FJSP-LB-MK}). The core challenge lies in balancing two objectives: optimizing the operation-to-machine assignment for compact scheduling, while consecutively scheduling jobs containing similar part types (i.e., steel plates) to maximize pallet utilization and reduce pallet switching caused by category mismatches.
\begin{figure}[t]	
	\centering
	\includegraphics[scale=0.22]{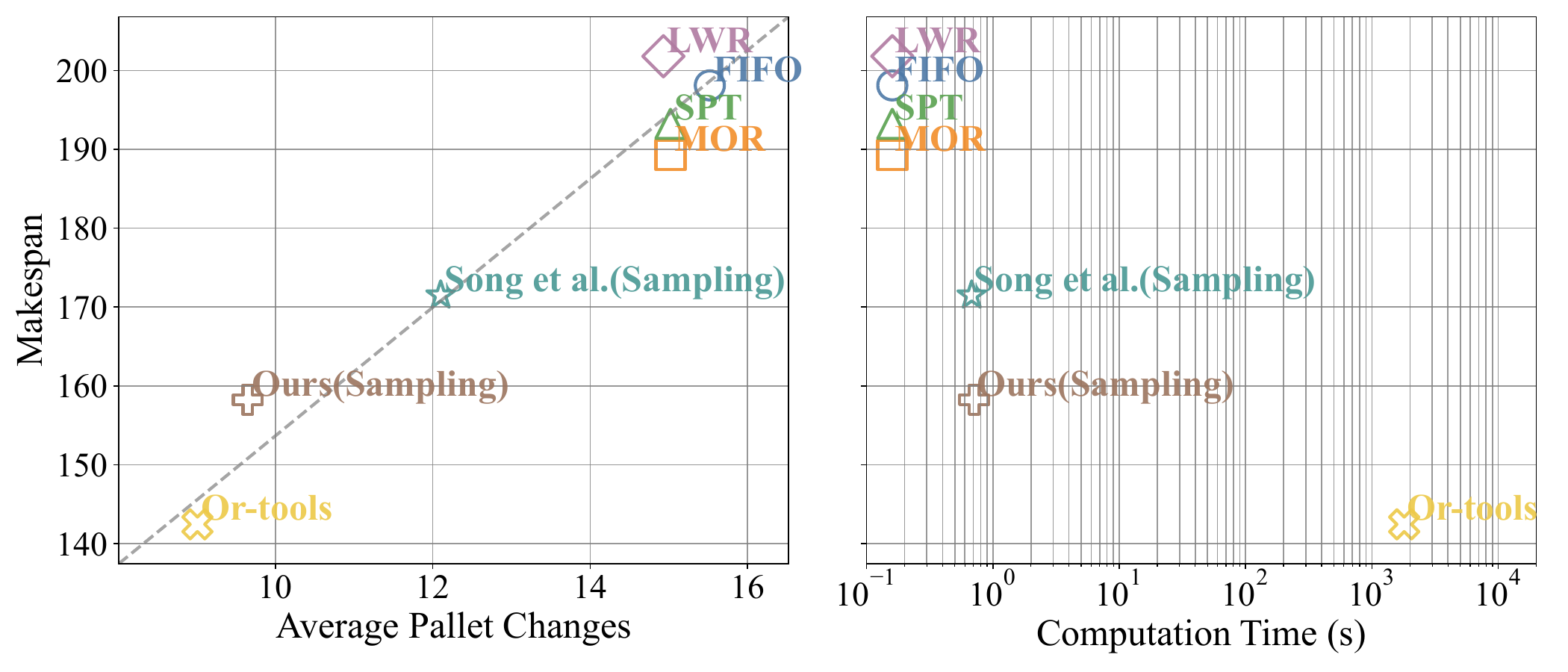}
	\caption{Comparison of the makespan, pallet changes, and computation time (the closer to the bottom and left, the better). Our method establishes an effective balance between solution quality and computational cost.}
	\label{fig_alg_compare}	
\end{figure}
\begin{figure*}[t]	
	\centering
	\includegraphics[scale=0.9]{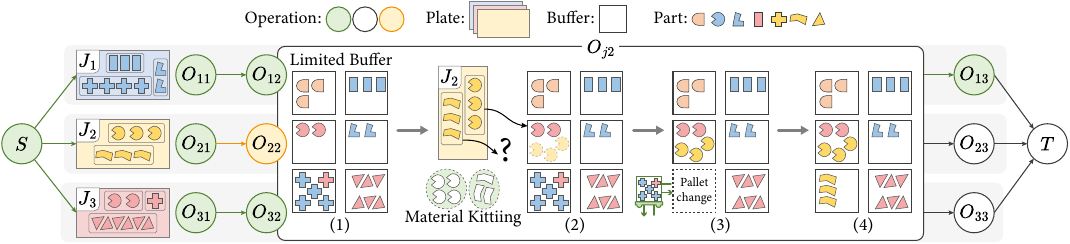}
	\caption{FJSP with limited buffer and material kitting constraints. Green, white, and yellow circles indicate scheduled, unscheduled, and currently being scheduled operations. Parts of the same job share color; identical shapes denote the same part type. Operation \( O_{22} \) is subject to limited buffer and kitting constraints: (1) Six pallets are available, each preloaded with parts; (2) Job \( J_2 \)'s parts must be split across two pallets, but one type is not in existing categories; (3) A pallet must be replaced with an empty one; (4) Once an empty pallet is available, all parts can be properly assigned.}
	\label{fig_problem_state}	
\end{figure*}

There are numerous methods for solving FJSP and its variants. Traditional methods, such as constraint programming, rely on centralized search methods, and the computational time cost becomes prohibitive as the problem scales. Heuristic and metaheuristic algorithms often rely on manually designed rules, which leads to a lack of generalization. In recent years, Deep Reinforcement Learning (DRL) has demonstrated considerable promise in solving the standard FJSP and a range of combinatorial optimization problems \cite{zhao2022learning,zhao2025deliberate,lin2025decoupled}. It supports end-to-end decision-making with second-level inference time and consistently competitive performance, owing to its long-term reward mechanism in sequential decision-making and its ability to autonomously learnb  effective heuristics for optimization tasks \cite{bello2016neural}. However, when faced with complex practical constraints in production line scenarios, existing DRL methods show significant limitations. Most of these methods rely on simplified state representations and struggle to capture long-term, non-local state dependencies arising from shared resources (e.g., pallets). As a result, the agent cannot anticipate how current decisions actually affect the availability of shared resources, leading to poor performance in scenarios under complex constraints.

To overcome these limitations, we build upon the strengths of DRL in long-sequence decision-making and leverage a heterogeneous graph network for global state modeling in the presence of buffer constraints. By constructing the dependency among different instances within the graph, we enable efficient message passing of critical information. During this process, we give higher attention to decisions that might lead to costly pallet changes, providing the decision network with reliable state feature embeddings. Ultimately, the model learns a scheduling preference that inherently incorporates material kitting logic, decomposing complex global constraints into learnable local signals on the graph, thereby enhancing state representation in complex dynamics and contributing to overall decision making.

Our main contributions are as follows:
\begin{enumerate}
    \item We are the first to address the FJSP problem under buffer constraints based on a DRL framework, demonstrating superior performance over baseline methods.
    \item 
    % To tackle the limitation of DRL methods in modeling complex global state representations, 
    We effectively leverage and enhance a heterogeneous graph neural network (HGNN), enabling a more precise construction of state dependencies, allowing the model to perceive the costs introduced by pallet change (switch) and thereby focus on high-cost operations. As a result, the network produces reliable state feature embeddings that lead to improved decision performance.
    \item 
    % Motivated by the goal of minimizing pallet changes to reduce makespan, 
    We validate the proposed methods and multiple baseline approaches through experiments on synthetic and real production-line datasets. Experimental results across these datasets show that our method achieves a favorable balance between performance and computational efficiency, as depicted in Figure~\ref{fig_alg_compare}.
    
    % that lead to increased makespan).
\end{enumerate}
\section{Related Works}

\subsection{Traditional Approaches for FJSP}
Traditional approaches to FJSP include exact methods, heuristics, and metaheuristics. Exact methods such as Mixed Integer Linear Programming (MILP) \cite{zhao2024batch} and Constraint Programming \cite{muller2022algorithm} can obtain optimal solutions for small-scale instances but become computationally prohibitive as problem size increases. Heuristic methods, such as Priority Dispatching Rules (PDRs) like FIFO \cite{chen2013flexible}, generate fast feasible solutions but suffer from myopic decisions and limited generalization due to manually designed rules. Metaheuristics, including Genetic Algorithms \cite{chen2020self} and Differential Evolution \cite{li2022hybrid}, search for high-quality solutions but often struggle to balance solution quality and computational efficiency.

\subsection{DRL Approaches for FJSP}
DRL improves solution efficiency for FJSP through end-to-end policy learning. For example, \cite{yuan2024solving} reduced action complexity using operation–machine pairs, while \cite{zhang2020learning} integrated Graph Neural Networks (GNNs) with DRL to enhance PDRs. \cite{song2022flexible} employed heterogeneous graphs with attention mechanisms for state encoding, and \cite{echeverria2023solving} combined multi-policy generation with Proximal Policy Optimization (PPO) to handle large-scale instances. However, existing DRL approaches still struggle to construct expressive state representations and adequately capture complex state dependencies.
\subsection{FJSP Problems Under Constraints}
Real-world FJSP often involves additional constraints. For example, \cite{chen2023multi} considered fixture constraints using a hybrid genetic algorithm. Limited buffer zones for work-in-progress storage have also been studied, as \cite{han2019flexible} optimized scheduling with a public buffer, and \cite{liang2019multi} proposed a hybrid differential evolution algorithm for multi-objective scheduling under buffer limitations. However, material kitting—a critical requirement in many production environments—remains largely unexplored. To fill this gap, we investigate the FJSP-LB-MK problem by jointly incorporating limited buffer and material kitting constraints into scheduling optimization.

\section{Problem Statement}
% 
% \textbf{FJSP-LB-MK} is a variant of the FJSP problem abstracted from real production lines. In a real production line, a batch of steel plates needs to be processed each time, and each steel plate corresponds to a job in the FJSP problem. We need to determine an optimal scheduling order for the steel plates and establish task assignment rules for the steel plates to different machines to minimize the total job completion time. Compared to the ideal FJSP problem, we consider practical constraints present in actual production lines, such as the limited number of buffer zones (Limited Buffer) and material kitting requirements (Material Kitting, MK).

% As shown in Figure 1, some workstations have a limited number of buffer zones (specifically in the form of pallets), and parts cut from steel plates need to be placed on the same pallet according to the same category (this is called material kitting, MK). Different categories of parts must be placed in different pallets. Since the number of pallets is limited, if the number of part categories exceeds the number of pallets, a pallet must be taken out of storage and replaced with an empty one (we do not consider pallet capacity in this problem).

The FJSP-LB-MK is a variant of the classical FJSP that incorporates a more practical buffer constraint.
An instance of the FJSP is defined by a set of jobs \( J = \{J_1, J_2, ..., J_n\} \), and a set of machines \( M = \{M_1, M_2, ..., M_m\} \). Each job \( J_i \in J \) concludes a set of operations \( O_i = \{O_{i1}, O_{i2}, ..., O_{il}\} \), and each operation \( O_{ij} \) can be processed on machines from a subset \( M_{ij} \subset M \). The time taken by machine \( M_k \) to process operation \( O_{ij} \) is denoted as \( p_{ijk} \in \mathbb{R}^+ \). The most common objective function used in the FJSP is the Makespan, denoted as $C_{max}$, which is defined as: 
\begin{equation}\label{eq_3_1}
	\begin{aligned}
		C_{\max}= \max C_{ij},
	\end{aligned}
\end{equation}
where \( C_{ij} \) represents the completion time of operation \( O_{ij} \) for job \( J_i \). In addition to the classical FJSP, the FJSP-LB-MK introduces two key constraints: limited buffer and material kitting, as shown in Figure \ref{fig_problem_state}. Limited buffer means there is a limited number of buffer zones (each zone contains a pallet) available to store parts from each steel plate (the “job” in FJSP-LB-MK).  Material kitting means each pallet can hold only parts of a single category (parts from different steel plates but having the same category can be grouped together).

Let \( B = \{B_1, B_2, ..., B_k\} \) be the set of \( k \)  buffer zones (pallets). The current total part categories in the buffer zone can be represented by \( P_b = [p_{1}, p_{2}, ..., p_{c}] \). For a newly arrived steel plate \( J_i \) at the part-sorting operation, the part categories of which is given by \( P_{J_i} = [p_{1}, p_{2}, ..., p_{J_i}] \), for parts in \( J_i \) belonging to existing categories of buffer zone, they are directly assigned to their corresponding pallets, for new part categories, they are assigned to the empty pallets. If new categories exceed the number of empty pallet, one or more pallets currently storing parts need to be moved to the warehouse and replaced with a new empty pallet to accommodate the new category of parts. The single pallet change time is defined as \(t_\text{switch} \), the number of exceeded new part categories is \(N_{\text{excess}}\), and since only one pallet can be changed at a time, the total pallet change time is, 
\begin{equation}\label{eq_3_2}
	\begin{aligned}
		T_{\text{replace}} = N_{\text{excess}} \times t_{\text{switch}}.
	\end{aligned}
\end{equation}

%% Very key !!!
%Each steel plate contains a different number of part categories, and there will be some overlap in the categories between different steel plates (there are intersections). Therefore, if multiple steel plates with more overlapping part categories are scheduled consecutively, more parts of the same category can be placed in the same pallet (material kitting for the same parts). This improves the pallet utilization rate, directly reduces the frequency of pallet switching, and thus reduces the consumption of processing time.

\section{Deep Reinforcement Learning for FJSP-LB-MK}

\subsection{MDP Formulation}

Under the DRL framework, the scheduling process is typically modeled as a Markov Decision Process (MDP). At each decision step, the RL agent receives the current global state and selects an operation-machine pair as the scheduling action. This action triggers a state transition, and the updated state becomes the input for the next decision. The process continues until all operations are scheduled.

\begin{figure}[t]	
	\centering
	\includegraphics[scale=0.66]{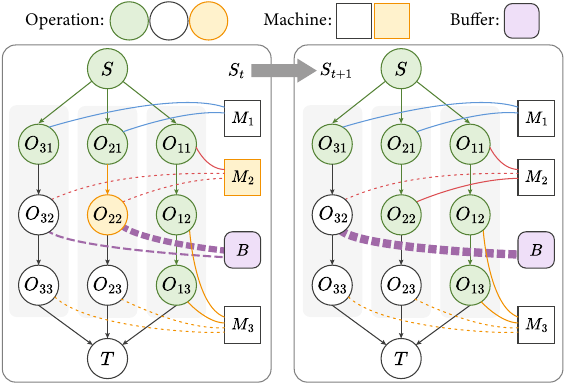}
	\caption{Example of state transition. At timestep \(t\), both unscheduled operations \(O_{32}\) and \(O_{22}\) are subject to buffer constraints; therefore, the buffer node is connected to both of them. The algorithm then prepares to execute the \(O_{22}-M_{2}\) action. Upon execution, the graph transitions to state \(s_{t+1}\). The newly scheduled operation node is marked green, and its associated connections are updated accordingly.}
	\label{fig_state_update}	
\end{figure}

\subsubsection{State}\label{state}  
As shown in Figure \ref{fig_state_update}, we use a heterogeneous graph \(\mathcal{H}\) to represent the global state. This directed graph consists of multiple node and edge types. For state feature design, we primarily follow \cite{song2022flexible}. For operation features, we adopt the 6-D scheduling features in \cite{song2022flexible}, and additionally introduce part category one-hot features \textbf{Type} \(\in \mathbb{R}^{T}\), where \( T \) denotes the number of part categories, to represent the part types associated with the job of the operation, a binary indicator \textbf{PS} \(\in \mathbb{R}\) to distinguish part-sorting operations, and \textbf{SwEst} \(\in \mathbb{R}\) to estimate the pallet change count based on the current state. For buffer features, we use a concatenated vector to represent the status of all buffers, which includes: 1) part category one-hot features \textbf{Type} \(\in \mathbb{R}^{T}\) to capture the part types stored in all buffers and 2) occupancy rate \(\in \mathbb{R}\) to indicate the overall utilization of the buffers. As the scheduling progresses, the connections of edges and associated state features of these nodes will dynamically change.

\subsubsection{Action}
Our approach integrates operation selection and machine assignment into a compound decision. At each decision point \(t\), the action space \(A_{t}\) consists of all eligible operation-machine pairs—where an operation \(O_{ij}\) is eligible once its predecessor completes and a machine \(M_{k}\) is idle. This action space shrinks over time. The set of operations in \(A(t)\) is called the candidate operation set, \(J_c(t)\).

\subsubsection{Transition}  
As illustrated in Figure~\ref{fig_state_update}, after executing action $a_t$, the environment transitions to a new state $s_{t+1}$, updating the relevant operation and machine sets. The completed operation node and its associated edges are removed, and the features and connections of machine, operation, and buffer nodes are refreshed to reflect the updated schedule. For part-sorting operations with buffer constraints, a dedicated execution process is triggered. Specifically, these operations enter a \emph{Pallet Change} module, which determines whether pallet changes are necessary and how many are required. 

\subsubsection{Reward}
Our reward function has two components: the estimated change in makespan and the change in pallet changes, both based on the differences between states \( s_t \) and \( s_{t+1} \). This dual-component reward guides the network toward learning a policy that minimizes makespan while taking into account pallet changes. The reward is defined as follows:
\begin{equation}\label{eq_reward}
	\begin{aligned}
		r(s_t, a_t, s_{t+1})= C_{\max}(s_t)-C_{\max}(s_{t+1}) 
        \\ +  \lambda (P_{\max}(s_t) - P_{\max}(s_{t+1})),
	\end{aligned}
\end{equation}
where $P_{\max}(s_t)$ represents the total pallet changes at time step \(t\), \(\lambda\) is a weight factor to link the two objectives. %Based on empirical tuning, we selected different $\lambda$ for the synthetic and production-line datasets.

\subsection{Heterogeneous GNN}
\begin{figure}[t]	
	\centering
	\includegraphics[scale=0.31]{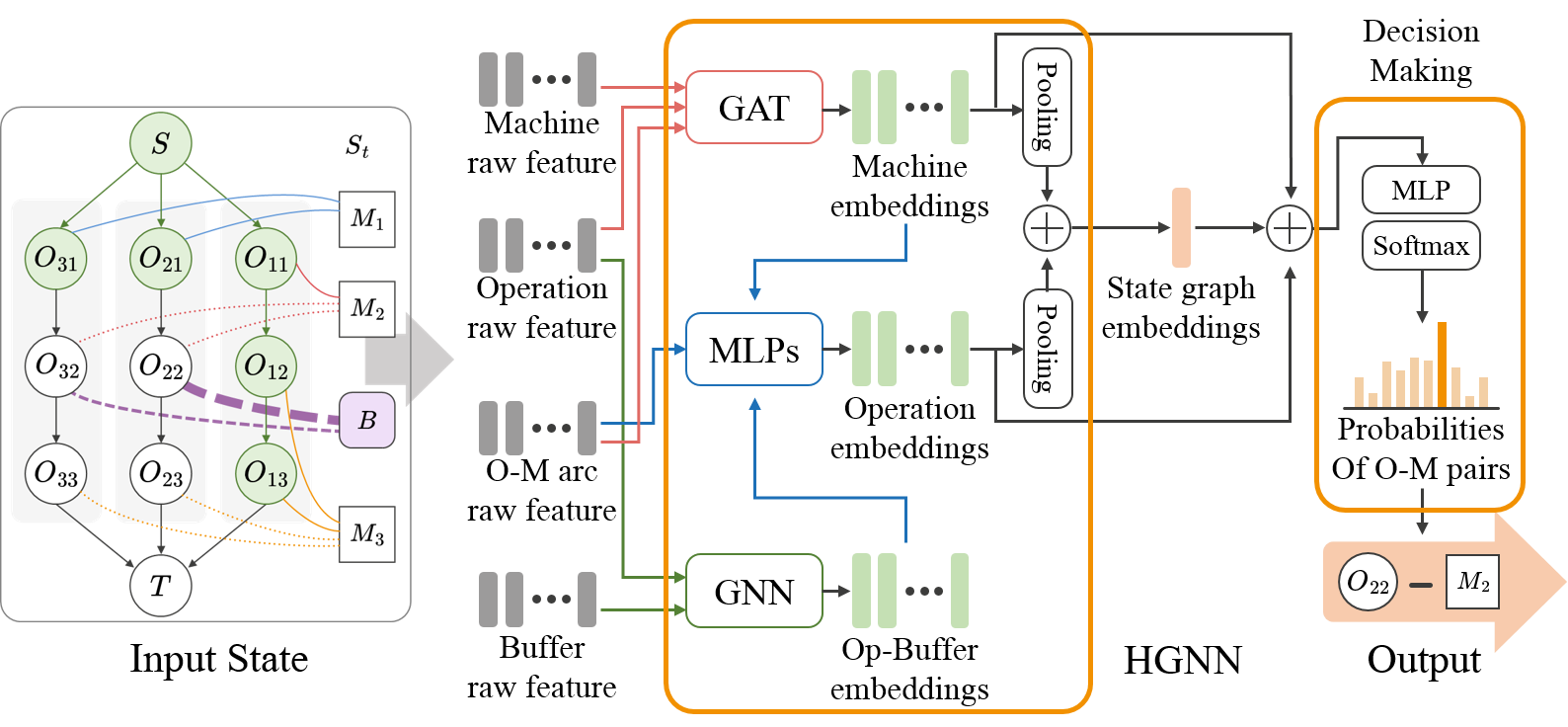}
	\caption{Heterogeneous GNN uses multiple state features as input, and outputs the O-M pair decision.
}
	\label{fig_Net_work}	
\end{figure}

To effectively extract and represent the global state information of FJSP-LB-MK, we draw on and enhance the HGNN proposed by \cite{song2022flexible}, and the enhanced network architecture is illustrated in Figure \ref{fig_Net_work}. Specifically, the enhanced HGNN comprises three main stages: Machine Embedding, Operation-Buffer Embedding, and Operation Embedding. For the Machine Embedding and Operation Embedding, we mainly follow the original setting of HGNN. The Operation-Buffer Embedding is a novel module we introduced to effectively propagate the current state information of buffers to operations,  and incorporate the new Operation-Buffer feature to the Operation Embedding.

\begin{figure}	
	\centering
	\includegraphics[scale=0.5]{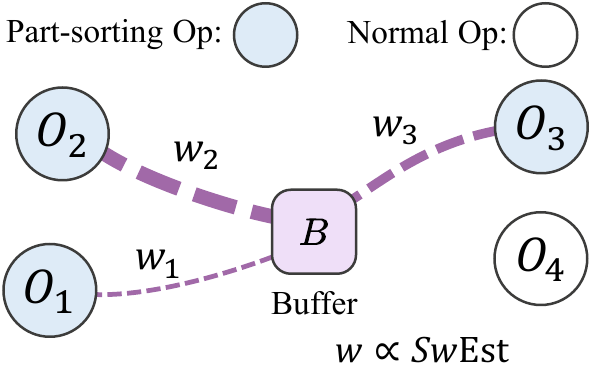}
	\caption{Message passing between the buffer and the part-sorting operations, \(w\) is the weight of the edge}
	\label{fig_message}	
\end{figure}

For Operation-Buffer Embedding, as shown in Fig.~\ref{fig_state_update}, the heterogeneous graph of FJSP-LB-MK incorporates a unique buffer node \(B\), connected exclusively to all part-sorting operation nodes to convey the buffer state information at each time step. To further inform the network of potential pallet change costs, we employ a weighted GNN message-passing mechanism, which propagates buffer features to its neighboring part-sorting operation nodes via weighted edges (Fig.~\ref{fig_message}). The edge weight is set directly proportional to the estimated number of pallet changes (switches) \textbf{SwEst} required for a specific operation \(O_{ij}\) at each time step: 
\begin{equation}
    \label{eq:op_buffer_weight}
    w_{ij} = \text{sigmoid}(\alpha \cdot SwEst),
\end{equation}
where \(\alpha\) is a scaling factor (0.3 in our setting). The Operation-Buffer feature of \(O_{ij}\in \mathcal{N}_t(B)\) is then computed as:
\begin{equation}
    \label{eq:op_buffer_feature}
    \delta_{ij} = w_{ij}B.
\end{equation}
In this way, operations with higher anticipated pallet change costs receive a more heavily weighted message from the buffer during feature aggregation.

 We highlight the key design choices in this component:

\subsubsection{Selective connectivity} \label{Selective connectivity} Instead of connecting the buffer node \(B\) to all operation nodes, we only connect it to the part-sorting operation nodes. This is because a GNN aggregates information from neighboring nodes, and indiscriminately broadcasting buffer features to all operations—regardless of their relevance—would not only introduce noise but also weaken the ability of the network model to identify which operations should attend to the pallet state. To demonstrate the advantages of Selective connectivity, we evaluate and compare various connectivity strategies between buffer and operation nodes in the ablation studies.

\subsubsection{Cost-sensitive propagation}\label{Cost-sensitive propagation} Instead of employing uniform message broadcasting, we dynamically modulate propagation through edge weights that are positively correlated with the estimated pallet change (switch) cost  (\textbf{SwEst}). This mechanism, referred to as the \textit{cost-avoiding} strategy, encourages the network model to prioritize decisions associated with higher switch costs. In contrast, we also investigate a \textit{benefit-seeking} strategy, where edge weights are negatively correlated with \textbf{SwEst}, thereby guiding the network model toward decisions with lower immediate switch costs. Our ablation study confirm the superiority of the \textit{cost-avoiding} strategy. We attribute this to the fact that benefit-seeking may lead to locally optimal but short-sighted choices, neglecting downstream constraints and creating bottlenecks. By contrast, cost-avoiding encourages the model to proactively mitigate high-cost decisions, achieving better global optimization and more robust scheduling performance.

\section{Experiments}
\subsection{Dataset}
To evaluate the performance of our proposed algorithm, we conducted a series of experiments on synthetic datasets of varying scales and real production line datasets. 
\subsubsection{Synthetic Dataset}
Similar to most studies on FJSP, synthetic FJSP-LB-MK instances were generated for model training and testing, with the generation procedure following the methodology outlined in \cite{brandimarte1993routing}. We considered six distinct problem scales, ranging from 10 jobs $\times$ 5 machines to 40 jobs $\times$ 10 machines. Taking into account the specific characteristics of the FJSP-LB-MK problem, we further customized these synthetic instances with modifications:
\begin{itemize}
  \item Within the set of operations for each job, certain operations were designated as \textbf{part-sorting} operations. These operations are only executable by specific machines.
  %, with their processing time calculated as: the number of parts $\times$ the time per part + the number of pallet switches $\times$ the time per pallet switch. 
  \item A subset of machines was designated exclusively for executing part-sorting operations. 
  \item For each problem scale, several variant instances were created based on the number of machines and operations.
\end{itemize}
%This design serves two purposes: 1) enhancing the diversity of problem instances to evaluate the algorithm's generalization under varying part types and buffer configurations; 2) mirroring the resource and storage disparities encountered in real production environments, laying the groundwork for future experiments with real-world datasets. 
As shown in Table \ref{tab:Synthetic dataset}, the first four columns represent the original FJSP parameters, while the last five columns correspond to the extended FJSP-LB-MK parameters, including part sorting, pallet change, and other features. These parameters were adaptively adjusted with the problem scale.
\begin{table}[t]
\centering
\caption{Parameters of the Synthetic FJSP-LB-MK Datasets}
\scriptsize
\setlength{\tabcolsep}{2.5pt} % 缩小列间距
\begin{tabular}{cccccccccc}
\toprule
\textbf{Size ($n \times m$)} & $n_i$\textsuperscript{1} & $|\mathcal{M}_{ij}|$\textsuperscript{2} & $\bar{p}_{ij}$\textsuperscript{3}
& $n_{ps}$\textsuperscript{4} & $c_j$\textsuperscript{5} & $C$\textsuperscript{6} & $t_p$\textsuperscript{7} & $t_r$\textsuperscript{8} & $P$\textsuperscript{9} \\
\midrule
$10 \times 5$  & U(4, 6)   & U(1, 5)   & U(1, 20)   & 1 & U(3, 5) & 10 & 2 & 5 & 6 \\
$20 \times 5$  & U(4, 6)   & U(1, 5)   & U(1, 20)   & 1 & U(3, 5) & 10 & 2 & 5 & 6 \\
$15 \times 10$ & U(8, 12)  & U(1, 10)  & U(1, 20)   & 1 & U(3, 5) & 10 & 2 & 5 & 6 \\
$20 \times 10$ & U(8, 12)  & U(1, 10)  & U(1, 20)   & 1 & U(3, 5) & 10 & 2 & 5 & 6 \\
$30 \times 10$ & U(8, 12)  & U(1, 10)  & U(1, 20)   & 1 & U(3, 5) & 10 & 2 & 5 & 6 \\
$40 \times 10$ & U(8, 12)  & U(1, 10)  & U(1, 20)   & 1 & U(3, 5) & 10 & 2 & 5 & 6 \\
\bottomrule
\end{tabular}

\vspace{1ex}

\begin{tabular}{@{}l@{}}
\textsuperscript{1} Number of operations in Job $J_i$; \textsuperscript{2} Number of compatible machines for operation \\ $O_{ij}$;
\textsuperscript{3} Average processing time of operation $O_{ij}$; 
\textsuperscript{4} Number of part-sorting opera- \\ tions per job; \textsuperscript{5} Number of part categories on each job; 
\textsuperscript{6} Total number of part cate- \\ gories; 
\textsuperscript{7} Part placement time (sec); 
\textsuperscript{8} Pallet replacement time	(sec);
\textsuperscript{9} Number of \\ available pallets.
\end{tabular}

\label{tab:Synthetic dataset}
\end{table}
\begin{table}[t]
\centering
\scriptsize
\setlength{\tabcolsep}{2.5pt}
\caption{Parameters of the Real Production Line Dataset}
\begin{tabular}{ccccccccc}
\toprule
\textbf{Size ($n \times m$)} & $n_i$ & $|\mathcal{M}_{ij}|$ 
& $n_{ps}$ & $c_j$ & $C$ &
$t_p$ & $t_r$ & $P$ \\
\midrule
A: $20 \times 16 $   & 8   & $\{5:1, 1:3, 2:4\}$  & 1 & 1-11 & 18--47 & 14 & 90 & 18\\
B: $20 \times 12 $   & 9   & $\{8:1, 1:4\}$   & 3 & 1--12 & 16--38 & 14 & 180 & 48\\
C: $20 \times 10 $  & 9  & $\{8:1, 1:2\}$   & 3 & 1--9 & 17--28 & 14 & 180 & 24\\
D: $20 \times 12 $  & 11 & $\{10:1, 1:2\}$   & 4 & 1--8 & 26--55 & 14 & 90 & 20\\
\bottomrule
\end{tabular}
%\vspace{1ex}
% \begin{tabular}{@{}l@{}}
% \textsuperscript{1} Number of operations in Job $J_i$; \textsuperscript{2} Number of compatible machines for operation \\ $O_{ij}$;
% \textsuperscript{3} Number of part-sorting operations per job; 
% \textsuperscript{4} Number of part categories on \\ each job; 
% \textsuperscript{5} Total number of part categories;
% \textsuperscript{6} Part placement time (sec); 
% \textsuperscript{7} Pallet \\ replacement time	(sec);
% \textsuperscript{8} Number of available pallets.\\
% \end{tabular}
\label{tab:Real-world dataset2}
\end{table}
\subsubsection{Real Production Line Dataset}
We constructed a real production-line dataset by collecting comprehensive data related to steel plate processing from four industrial production lines. In these production lines, each steel plate (job) $J$ undergoes a sequence of operations. For each operation $O_i$, the factory information system automatically records processing-related information. Combined with a dedicated processing-time calculation program, this enables the computation of the required processing time $p_{ik}$ when the operation is executed on an eligible machine $M_k$. For each production line, we sampled 10,000 instances for training, 100 for validation, and 100 for testing. Each instance represents a production segment consisting of 20 jobs (steel plates). The key parameters for the four production lines are shown in Tables \ref{tab:Real-world dataset2}. Unlike the uniform sampling of the synthetic dataset, the $|\mathcal{M}_{ij}|$ metric of the dataset reflects actual machine-operation mappings. 
% For example, \{5:1, 1:3, 2:4\}\ in Table \ref{tab:Real-world dataset2} represents the following:
% \begin{itemize}
%     \item 5 operations can be performed by a specific machine;
%     \item 1 operation can be performed by 3 machines;
%     \item 2 operations can each be performed by 4 machines.
% \end{itemize}

Additionally, part categories and their counts are derived from production lines, providing actual values for $c_j$ and $C$, resulting in greater variability than in the synthetic dataset, and making the production line dataset a more rigorous benchmark for testing an algorithm’s adaptability to buffer congestion. In this work, we idealize the production environment by not accounting for real-world disruptions such as machine failures or dynamic order insertions.

\subsection{Implementation Details}
\begin{algorithm}
\caption{Details of Training Procedure with PPO}
\label{alg:ppo_training_short}
{\small
\setlength{\baselineskip}{8.5pt}
\begin{algorithmic}[1]
\setlength{\itemsep}{0pt}
\setlength{\parsep}{0pt}
\setlength{\parskip}{0pt}
\REQUIRE Initial heterogeneous GNN, policy network (actor $\pi_\theta$), and value network (critic $v_\phi$); total iterations $I$
\STATE Sample $B$ instances
\FOR{$\text{iter} = 1$ to $I$}
    \FORALL{$b = 1$ to $B$ \textbf{in parallel}}
        \STATE Set initial state $s_t$ for instance $b$
        \WHILE{$s_t$ not terminal}
            \STATE Get embeddings from network
            \STATE Select $a_t \sim \pi_\theta(s_t)$; observe $r_t$, $s_{t+1}$
            \STATE $s_t \leftarrow s_{t+1}$
        \ENDWHILE
        \STATE Estimate advantage $A_t$ and compute loss $\mathcal{L}$
        \STATE Update $\theta$, $\phi$, $\omega$ for $R$ epochs
    \ENDFOR
    \IF{$\text{iter} \bmod 10 = 0$}
        \STATE Validate policy
    \ENDIF
    \IF{$\text{iter} \bmod 20 = 0$}
        \STATE Resample instances
    \ENDIF
\ENDFOR
\RETURN Updated model parameters.
\end{algorithmic}}
\end{algorithm}
\begin{table}[t]
\caption{Training parameters.}
\centering
\begin{tabular}{|p{5cm}|p{1.7cm}|}
\hline
\textbf{Parameter} & \textbf{Value} \\
\hline
lr (learning rate) & 2e-4 \\
\hline
gamma (discounting factors) & 1.0 \\
\hline
K\_epochs (update epoch in PPO) & 3 \\
\hline
A\_coeff (weight of policy loss) & 1 \\
\hline
vf\_coeff (weight of value loss) & 0.5 \\
\hline
entro\_coeff (weight of entropy loss) & 0.05 \\
\hline
KL\_coeff (weight of KL regularization term) & 0.05 \\
\hline
parallel\_iter (batch size of training)& 20 \\
\hline
save\_timestep (validate interval) & 10 \\
\hline
max\_iterations & 10000 \\
\hline
minibatch (update batchsize in PPO) & 512 \\
\hline
update\_timestep (update interval)& 5 \\
\hline
\end{tabular}
\label{tab: training parameters}
\end{table}

\subsubsection{Details of training} All training and experiments were performed on a device equipped with an NVIDIA GeForce RTX 3080 Ti GPU and an Intel Core i9-10980XE CPU. We use Proximal Policy Optimization (PPO) for training, which employs an actor-critic framework. The actor is the policy network \(\pi_{\omega}\), while the critic \( v_{\phi} \) predicts the value \( v(s_t) \) of a state \( s_t \). 
The training process involves \( I \) iterations, where a batch of instances is processed in parallel by the DRL agent with instance replacement every 20 iterations. Additionally, the policy is validated on a set of independent validation instances every 10 iterations during training. The training procedure and some key training parameters are shown in Algorithm \ref{alg:ppo_training_short} and Table \ref{tab: training parameters}.

\subsubsection{Training on datasets} First, we train our model on small-scale synthetic instances with 10 jobs $\times$ 5 machines, with the training instances generated on-the-fly, and we apply the policy model learned from the small-scale to all scales of the synthetic test set to verify the performance. 

For the real production line datasets, we use the model trained on the synthetic dataset as the pretrained model, and then fine-tune it on different production line datasets. Initially, when directly transferring the model from the synthetic dataset to the production line datasets, we observed a temporary deterioration in the makespan metric. To address this issue, we incorporated an additional KL-divergence regularization term in the PPO loss with respect to the pre-trained policy, which prevents overly rapid policy updates and leads to more stable performance improvements during training.
%For testing, we sample 100 instances for each problem scale.

%The architecture and parameter details of the network model are presented in the Appendix.

% \subsubsection{Hardware} All training and experiments are conducted on a machine equipped with an NVIDIA 3080Ti GPU and an Intel Core i9-10980XE CPU.
\subsection{Baseline}  
We adapt several strong-performing algorithms from the standard FJSP to the FJSP-LB-MK setting. These include: 1) the CP-SAT solver from Google OR-Tools, 2) Priority Dispatching Rules (PDRs), and 3) a DRL method known for its effectiveness on FJSP.

\textbf{OR-Tools.}  
OR-Tools is a widely used constraint programming solver \cite{ortools2024}. We use it with part sorting sequences obtained via MCTS (10,000 simulations) as static input. Each instance is solved within a 1800s time limit, and the best solution found serves as an upper bound for evaluation.

\textbf{PDRs.}  
The scheduling process is split into two stages: operation sequencing and machine assignment. For sequencing, we use rules such as Most Work Remaining (MWR), Least Work Remaining (LWR), and First in First Out (FIFO). For machine assignment, we use the Earliest End Time (EET).

\textbf{DRL.}  
We adapt the DRL framework from Song et al.~\cite{song2022flexible}, which performs well on standard FJSP benchmarks, to the FJSP-LB-MK. Two strategies are tested: 1) \textbf{DRL-Greedy}: selects the highest probability action, and 2) \textbf{DRL-Sampling}: samples actions based on the probability distribution. Each instance is run 100 times, reporting the best result.

\subsection{Evaluation Metrics}
To assess both the efficiency and quality of the proposed method, we adopt the following evaluation metrics commonly used in FJSP literature:

\begin{itemize}
    \item \textbf{Makespan}: The average total completion time across all test instances, as the primary optimization objective.
    \item \textbf{Gap}: The average relative difference in makespan compared to the solutions obtained by OR-Tools, used to evaluate the quality of the solution.
    \item \textbf{Time}: The average computational time required to obtain solutions, reflecting the algorithm's efficiency.
\end{itemize}
In addition to the standard scheduling metrics, we introduce \textbf{Switches} as an indicator of the algorithm’s performance with respect to pallet exchange frequency.
\begin{itemize}
    \item \textbf{Switches}: The average total number of pallet changes across all test instances.
\end{itemize}
\subsection{Results and Analysis}
\begin{figure}[t]
	\centering
        \scriptsize
        \subfigure[\cite{song2022flexible}, Pallet\_change=16, Makespan=192]{
		\includegraphics[scale=0.25]{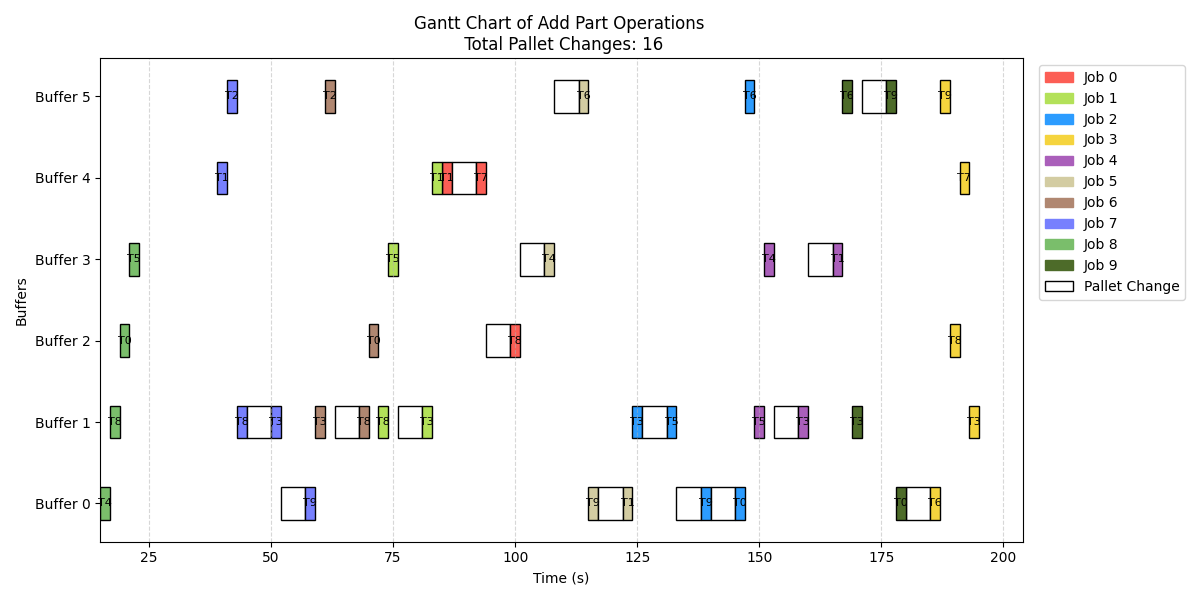}
		%
		%		\centerline{(b) 3d-3d error}
		\label{fig_pallet_change:a}		
	}
        \subfigure[Ours, Pallet\_change=10, Makespan=167]{
		\includegraphics[scale=0.25]{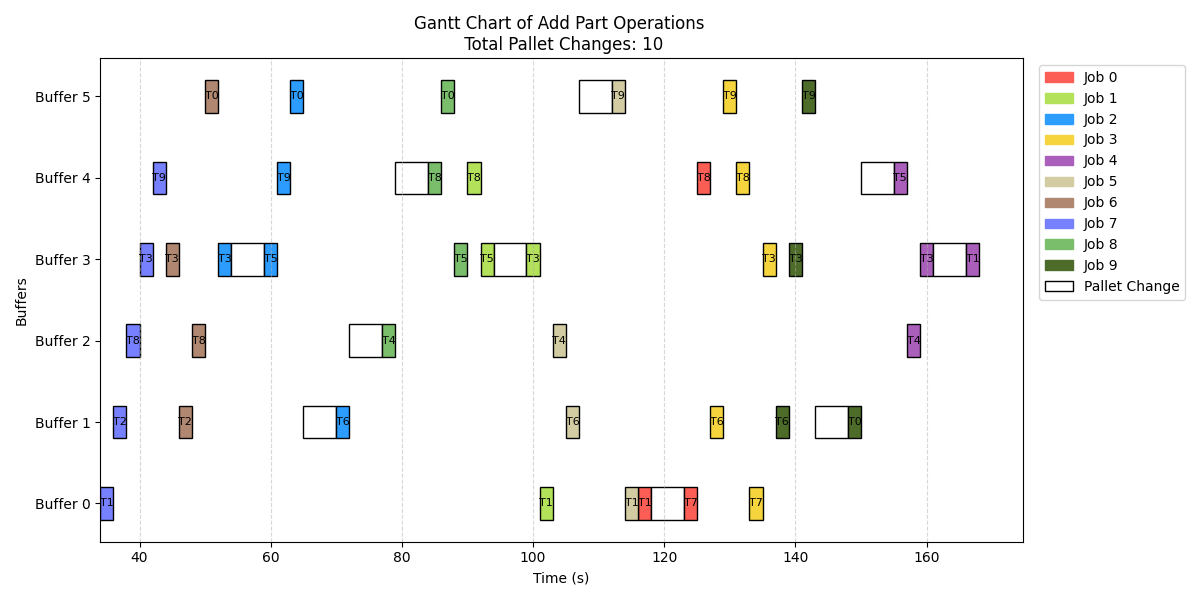}
		%
		%		\centerline{(b) 3d-3d error}
		\label{fig_pallet_change:b}		
	} 
	%	\captionsetup{justification=centering}
	\caption{Gantt Chart Comparison (Test Case: 10j5m test\_set instance). The white rectangles represent time delays caused by pallet changes. Rectangles of other colors represent the time intervals when parts from different jobs are loaded onto the pallet (where $T_{n}$ denotes the $n$-th type of part).} % The red, blue, and green curves show the details of \textbf{Ours}, \textbf{No\_object\_encoder}, and \textbf{No\_arm\_encoder} respectively.
	\label{fig_pallet_change}
\end{figure}
\begin{table*}[ht!]
	\caption{Performance on Synthetic FJSP-LB-MK Datasets.}
	\label{tab:syn}
	\centering
	\scriptsize
    \renewcommand{\arraystretch}{0.9}  % Reduce row height
	\begin{threeparttable}
		\begin{tabular}{cc|ccccc|cc|cc|c}
			\toprule[1pt]
			&       & \multicolumn{5}{c|}{PDRs}     & \multicolumn{2}{c|}{Greedy strategy} & \multicolumn{2}{c|}{Sampling strategy} & \\
			Size  &       & FIFO  & MOR & SPT   & MWR  & LWR  & \cite{song2022flexible}  & Ours  & \cite{song2022flexible}  & Ours  & OR-Tools\tnote{1} \\
			\midrule\midrule

			\multirow{4}[1]{*}{$10 \times 5$} 
			& Makespan & 198.11  & 189.27  & 193.23  & 191.74  & 201.81  & 177.02  & 169.85 & 171.48  & \textbf{158.25}\tnote{*} & \multirow{4}[0]{*}{142.45 (9.01)}   \\
			& Gap\tnote{2}      & 39.1\% & 32.9\% & 35.6\% & 34.6\% & 41.7\% & 24.3\%  & 19.2\% & 20.4\%  & \textbf{11.1\%} &   \\
			& Time (s) & 0.16    & 0.16    & 0.16    & 0.16    & 0.16    & 0.39   & 0.43            & 0.68    & 0.71            &   \\
			& Switches & 15.52       & 15.02       &  15.02       & 15.08       & 14.93      & 14.50       & 12.10       & 12.01      & \textbf{9.65}       &      \\
			\cmidrule{1-12}

			\multirow{4}[0]{*}{$20 \times 5$} 
			& Makespan & 359.06 & 373.83  & 361.26  & 373.95  & 373.54  & 342.69  & 300.70 &327.17  & \textbf{296.88} & \multirow{4}[0]{*}{279.45 (20.68)}  \\
			& Gap      & 28.5\% & 33.5\% & 29.3\% & 33.8\% & 33.7\% & 22.6\% & 7.6\% & 17.1\% & \textbf{6.2\%} &   \\
			& Time (s) & 0.32    & 0.32    & 0.32    & 0.32    & 0.32    & 0.78    & 0.84            & 1.30    & 1.41            &   \\
			& Switches & 32.21       & 32.86       & 33.01       & 33.18       & 33.50       & 32.40       & 21.61       & 27.66      & \textbf{19.19}       &      \\
			\cmidrule{1-12}

			\multirow{4}[0]{*}{$15 \times 10$} 
			& Makespan & 335.05 & 306.45  & 327.73  & 298.91  & 352.88  & 326.25  & 279.47 & 294.91  & \textbf{259.34} & \multirow{4}[0]{*}{239.79 (13.90)}  \\
			& Gap      & 39.7\% & 27.8\% & 36.7\% & 23.8\% & 47.2\% & 36.1\% & 16.5\% & 23.0\% & \textbf{8.2\%} &   \\
			& Time (s) & 0.51    & 0.51    & 0.50    & 0.50    & 0.50    & 1.17   & 1.33           & 1.91    & 2.02            &  \\
			& Switches & 24.46       & 24.07       & 24.24      & 24.21       & 23.83       & 23.62      & 16.03       & 20.76       & \textbf{13.79}       &      \\
			\cmidrule{1-12}

			\multirow{4}[1]{*}{$20 \times 10$} 
			& Makespan & 413.55  & 393.96  & 420.79  & 388.03  & 452.13  & 411.05 & 342.13 & 385.22  & \textbf{324.63} & \multirow{4}[0]{*}{297.75 (18.99)} \\
			& Gap      & 38.9\% & 32.3\% & 41.3\% & 30.3\% & 51.8\% & 38.1\%  & 14.9\% & 29.4\%  & \textbf{9.0\%} &   \\
			& Time (s) & 0.71    & 0.71    & 0.71    & 0.71    & 0.71    & 1.68   & 1.76            & 3.02    & 3.20            &   \\
			& Switches & 33.66      & 33.14       & 33.53      & 33.34       & 33.90       & 33.17       & 19.89       & 29.07       & \textbf{17.26}       &     \\
			\cmidrule{1-12}

			\multirow{4}[1]{*}{$30 \times 10$} 
			& Makespan & 595.15  & 566.64  & 593.82  & 554.79  & 624.53  & 579.29  & 452.11 & 547.34  & \textbf{444.70} & \multirow{4}[0]{*}{410.74 (25.93)}\\
			& Gap      & 44.9\% & 38.0\% & 44.6\% & 35.1\% & 52.0\% & 41.0\%  & 10.1\% & 33.3\%  & \textbf{8.3\%} &    \\
			& Time (s) & 1.25    & 1.25    & 1.25    & 1.25    & 1.25    & 2.26    & 2.64            & 6.82    & 7.19           &   \\
			& Switches & 51.98     & 51.05      & 52.26      & 51.73     & 51.13     & 50.57      & 26.31       & 45.23      & \textbf{25.13}       &      \\
			\cmidrule{1-12}

			\multirow{4}[1]{*}{$40 \times 10$} 
			& Makespan & 757.67  & 741.12  & 762.47 & 723.96  & 794.56  & 742.75  & 567.47 & 715.23  & \textbf{564.89} & \multirow{4}[0]{*}{518.18 (41.82)} \\
			& Gap      & 46.2\% & 43.0\% & 47.1\% & 39.7\% & 53.3\% & 43.3\%  & 9.5\% & 38.0\%  & \textbf{9.0\%} &   \\
			& Time (s) & 2.10    & 2.10    & 2.10    & 2.10    & 2.10    & 3.21    & 3.47           & 13.18    &   14.05         &   \\
			& Switches & 70.12      & 70.02      & 68.89      & 69.56      & 68.33     & 67.66      & 33.76      & 62.73      & \textbf{31.47}      &     \\
			\cmidrule{1-12}
		\end{tabular}
        \begin{tablenotes}
			\scriptsize
            \item[*] Best among all methods excluding OR-Tools;\item[1] For OR-Tools, the makespan and the switches (in brackets) of optimally solved instances within a 1800s time limit are reported; \item[2] Gap is calculated based on the OR-Tools result as a reference.

		\end{tablenotes}
	\end{threeparttable}
\end{table*}
\begin{table*}[ht!]
	\caption{Performance on Real Production Line Datasets.}
	\label{tab:real}
	\centering
	\scriptsize
    \renewcommand{\arraystretch}{0.9}  % Reduce row height
	\begin{threeparttable}
		\begin{tabular}{cc|ccccc|cc|cc|c}
			\toprule[1pt]
			&       & \multicolumn{5}{c|}{PDRs}     & \multicolumn{2}{c|}{Greedy strategy} & \multicolumn{2}{c|}{Sampling strategy} & \\
			Dataset  &       & FIFO  & MOR & SPT   & MWR  & LWR  & \cite{song2022flexible}  & Ours  & \cite{song2022flexible}  & Ours  & OR-Tools \\
			\midrule\midrule

			\multirow{4}[1]{*}{A} 
			& Makespan & 8946.92  & 8946.92  & 9188.00  & 8618.30  & 10413.75  & 8057.74  & 7753.52 & 7656.92  & \textbf{7392.72} & \multirow{4}[0]{*}{6706.06 (18.85)}   \\
			& Gap      & 33.4\% & 33.4\% & 37.0\% & 28.5\% & 55.3\% & 20.2\%  & 15.7\% & 14.2\%  & \textbf{10.2\%} &   \\
			& Time (s) & 0.92    &0.92     & 0.92    & 0.92    & 0.92    & 2.17   & 2.31            & 4.57    & 4.77          &   \\
			& Switches & 20.36       & 20.36       &  21.93       & \textbf{15.28}       & 27.12      & 16.36       & 18.34       & 15.86      & 16.42       &      \\
			\cmidrule{1-12}

			\multirow{4}[0]{*}{B} 
			& Makespan & 11867.21 & 11867.21  & 12731.46  & 12425.45  & 12902.02  & 11372.99  & 10855.39 & 10737.66  & \textbf{10207.62} & \multirow{4}[0]{*}{9949.58 (3.33)}  \\
			& Gap      & 19.3\% & 19.3\% & 28.0\% & 24.9\% & 29.7\% & 14.3\% & 9.1\% &7.9\% & \textbf{2.6\%} &   \\
			& Time (s) & 1.09    & 1.09    & 1.09    & 1.09    & 1.09    & 2.65    & 2.71            & 5.31    & 6.16            &   \\
			& Switches & 3.26       & 3.26       & 3.24       & 3.30       & 3.23       & 3.00       & 3.08       & 3.04     & \textbf{2.88}       &      \\
			\cmidrule{1-12}

			\multirow{4}[0]{*}{C} 
			& Makespan & 42896.40 & 42896.40 & 43803.81  & 40565.81  & 45554.45 & 41364.82  & 40460.43 & 40994.28  & \textbf{40359.80} & \multirow{4}[0]{*}{40377.03 (25.42)}  \\
			& Gap      & 6.2\% & 6.2\% & 8.5\% & 0.5\% & 12.8\% & 2.5\% & 0.2\% & 1.5\% & \textbf{-0.04\%} &   \\
			& Time (s) & 0.90    & 0.90    & 0.90   & 0.90    &  0.90   & 2.31   & 2.38         & 3.99    & 4.94            &  \\
			& Switches & 32.13      & 32.13       & 31.07      & 31.60       & 30.65       & 31.62      & \textbf{25.11}       & 25.63       & 25.89       &      \\
			\cmidrule{1-12}

			\multirow{4}[0]{*}{D} 
			& Makespan & 19801.34 & 19801.34  & 20502.43  & 20162.80  & 20535.88  & 18753.75  & 18475.30 & 18310.58  & \textbf{17616.80} & \multirow{4}[0]{*}{17074.17 (19.58)}  \\
			& Gap      & 16.0\% & 16.0\% & 20.1\% & 18.1\% & 20.3\% & 9.8\% & 8.2\% & 7.2\% & \textbf{3.2\%} &   \\
			& Time (s) & 1.30    & 1.30    & 1.30    & 1.30    & 1.30   & 2.88   & 2.97          & 5.26   & 6.74            &  \\
			& Switches & 27.81       & 27.81       & 28.17      & 27.08       & 28.37       & \textbf{19.17}     & 19.58       & 19.27      & 19.42       &      \\
			\cmidrule{1-12}
		\end{tabular}
	\end{threeparttable}
\end{table*}
Table~\ref{tab:syn} and Table~\ref{tab:real} report the scheduling performance of our method compared with various baselines on both the synthetic datasets and the real production line datasets. As shown, our approach consistently achieves superior makespan performance across all problem sizes and production lines. While our method requires more computation time than PDRs, it yields significantly better performance, and on the production line C dataset, it even surpasses the exact solver (OR-Tools) with markedly lower computation time.  Under both the \textit{Greedy} and \textit{Sampling} strategies, our method outperforms the DRL baseline~\cite{song2022flexible}, and as the problem scale grows, the performance gap continues to widen, demonstrating the superior scalability and generalization ability of our approach for large-scale problems.

More notably, in terms of the pallet changes (switches), our approach consistently outperforms the DRL baseline and all PDRs across all synthetic problem scales. This advantage can be attributed to the explicit incorporation of switch-related objectives into our model, where graph-based structural modeling and reward design guide the policy to reduce long-term pallet switches overhead.  However, on the production line datasets, our method does not always achieve the lowest number of switches. This phenomenon can be attributed to the significant heterogeneity across different production lines. Specifically, the total number of part categories $C$, the number of part categories per job $c_{j}$, and the available pallets $P$ vary substantially between lines, leading to disparate levels of buffer congestion. Consequently, while our strategy---pre-trained on synthetic instances and fine-tuned for production lines---primarily prioritizes makespan optimization, it may incur slightly more switches on certain real-world datasets. This reflects an inherent trade-off dictated by the specific buffer constraints of individual production environments.

Additionally, Figure \ref{fig_pallet_change} presents a comparative visualization of Gantt charts between our method and \cite{song2022flexible}, using a representative case from the 10j5m synthetic test set. The results show that our method can shorten the overall completion time by effectively reducing the number of pallet changes.

In summary, our method achieves dual optimization of scheduling quality and buffer usage cost, and achieves a good balance between solution quality and computational cost. This makes it particularly well-suited for large-scale flexible job-shop scenarios with practical buffer constraints.
 %% Evaluation on the Synthetic datasets.
% \input{AnonymousSubmission/LaTeX/Table/Table_synthetic_result_3}

% \input{AnonymousSubmission/LaTeX/Table/Table_synthetic_result_0724_Switch}

% \input{AnonymousSubmission/LaTeX/Table/Table_real_data_result_0726}
\begin{table}
	\caption{Effectiveness of Key State Features.}
	\label{tab:ablation_1}
	\centering
	\scriptsize
	\setlength{\tabcolsep}{0.4em}
    \renewcommand{\arraystretch}{0.6}  % Reduce row height
	\begin{threeparttable}
		\begin{tabular}{c|cc|cc}
			\toprule[1pt]
			     & \multicolumn{2}{c|}{Makespan}     & \multicolumn{2}{c}{Switches} \\
			  & Value  & Gap\_to\_ours & Value   & Gap\_to\_ours  \\
			\midrule

			PS+SwEst & 183.94 & 8.3\%  & 15.41  & 27.4\%  \\

			PS+Type & 187.73 & 10.5\%  & 15.42  & 27.4\%  \\

			Type+SwEst & 172.41 & 1.5\%  & \textbf{12.08}  & -0.2\%  \\
            
			\textbf{PS+Type+SwEst (Ours)} & \textbf{169.85} & 0.0\%  & 12.10  & 0.0\%  \\

			\toprule[1pt]
		\end{tabular}
	\end{threeparttable}
\end{table}

\begin{table}[t]
	\caption{Effectiveness of Selective Connectivity and Cost-sensitive Propagation.}
	\label{tab:ablation_2}
	\centering
	\scriptsize
	\setlength{\tabcolsep}{0.38em}
    \renewcommand{\arraystretch}{0.6}  % Reduce row height
	\begin{threeparttable}
		\begin{tabular}{c|cc|cc}
			\toprule[1pt]
			     & \multicolumn{2}{c|}{Makespan}     & \multicolumn{2}{c}{Switches} \\
			  & Value  & Gap\_to\_ours & Value   & Gap\_to\_ours  \\
			\midrule

			Base & 172.62 & 1.6\%  & 12.18  & 0.7\%  \\

			Pallet\_AllOps & 172.80 & 1.7\%  & 12.22  & 1.0\%  \\

			Pallet\_SortOnly & 171.03 & 0.7\%  & 12.31  & 1.7\%  \\
            
			Pallet\_SortOnly\_InverseWeight & 170.82 & 0.6\%  & 12.28  & 1.5\%  \\
            
			\textbf{Pallet\_SortOnly\_Weighted (Ours)} & \textbf{169.85} & 0.0\%  & \textbf{12.10}  & 0.0\%  \\

			\bottomrule[1pt]
		\end{tabular}
	\end{threeparttable}
\end{table}

\subsection{Ablation Study}
\begin{figure}[t]
	\centering
        \subfigure[Makespan]{
		\includegraphics[scale=0.26]{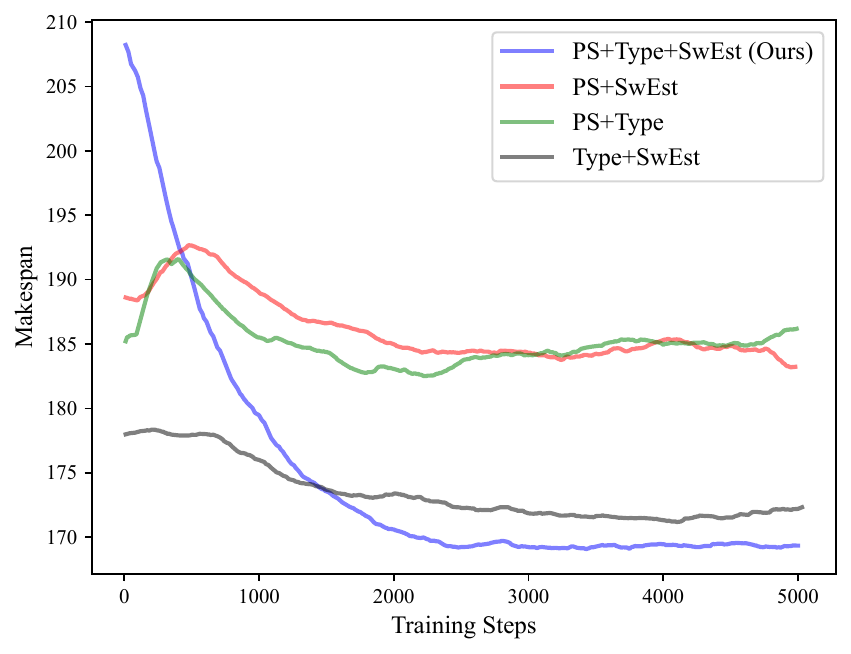}
		%
		%		\centerline{(b) 3d-3d error}
		\label{fig_delay:c}		
	}
        \subfigure[Switches]{
		\includegraphics[scale=0.26]{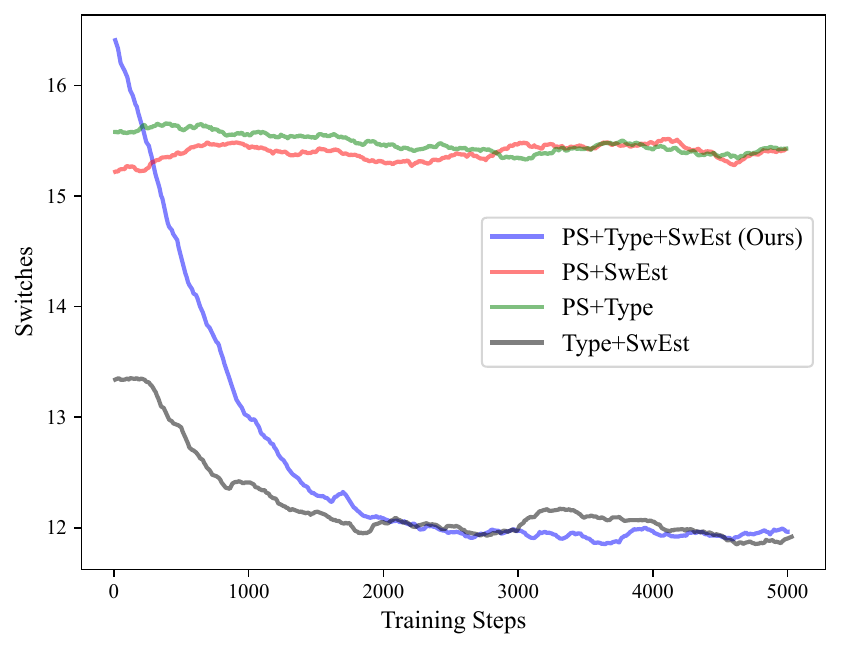}
		%
		%		\centerline{(b) 3d-3d error}
		\label{fig_delay:d}		
	} 
	%	\captionsetup{justification=centering}
	\caption{Training curve comparison on State Features.} % The red, blue, and green curves show the details of \textbf{Ours}, \textbf{No\_object\_encoder}, and \textbf{No\_arm\_encoder} respectively.
	\label{fig_ablation1}
\end{figure}
\begin{figure}[t]
	\centering
        \subfigure[Makespan]{
		\includegraphics[scale=0.26]{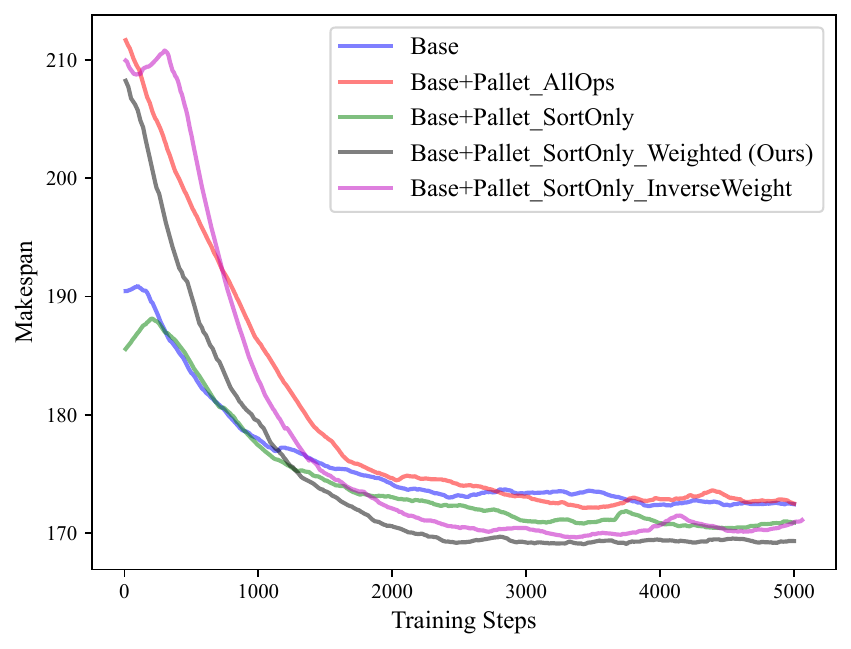}
		%
		%		\centerline{(b) 3d-3d error}
		\label{fig_delay:c}		
	}
        \subfigure[Switches]{
		\includegraphics[scale=0.26]{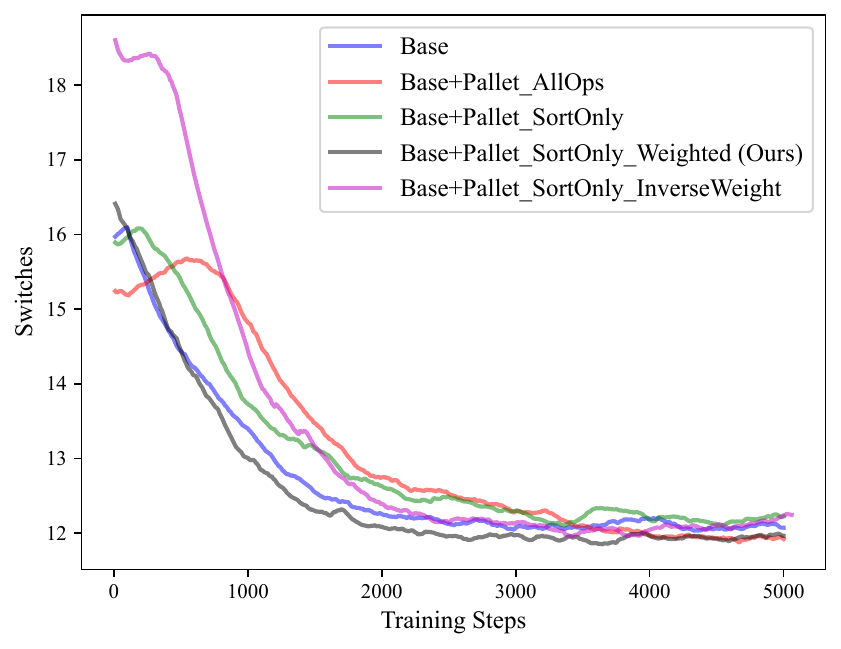}
		%
		%		\centerline{(b) 3d-3d error}
		\label{fig_delay:d}		
	} 
	%	\captionsetup{justification=centering}
	\caption{Training curve comparison on selective connectivity and cost-sensitive propagation.} % The red, blue, and green curves show the details of \textbf{Ours}, \textbf{No\_object\_encoder}, and \textbf{No\_arm\_encoder} respectively.
	\label{fig_ablation2}
\end{figure}

We perform ablation experiments on the smallest problem scale of the synthetic dataset (10 jobs $\times$ 5 machines), keeping all configurations consistent.

\subsubsection{Effectiveness of Key State Features}
To evaluate the impact of different input features, we remove components from the operation features. As mentioned in ~\ref{state}, the key features include: (1) 6-dimensional scheduling-related features used in \cite{song2022flexible}, (2) a binary indicator for part-sorting operations (\textbf{PS}), (3) a one-hot encoding of part types (\textbf{Type}), and (4) estimated pallet switches (\textbf{SwEst}). Figure \ref{fig_ablation1} and Table \ref{tab:ablation_1} show the results. Removing \textbf{Type} and \textbf{SwEst} significantly increases both \textit{Makespan} and \textit{Switches} (8.3\%/10.5\% and 27.4\%/27.4\% increases, respectively), highlighting their importance. Removing \textbf{PS} has a minor effect, suggesting its limited contribution in isolation. This emphasizes that \textbf{Type} and \textbf{SwEst} are crucial for estimating switch-related costs, while \textbf{PS} mainly helps identify operation types.

\subsubsection{Effectiveness of Selective Connectivity and Cost-sensitive Propagation}
In \ref{Selective connectivity} and \ref{Cost-sensitive propagation}, we introduce the selective connectivity and cost-sensitive propagation strategy. Figure \ref{fig_ablation2} and Table \ref{tab:ablation_2} show results from different configurations: \textbf{Base} means no connections between operation nodes and the buffer node, \textbf{Pallet\_AllOps} indicates all operations connected to the buffer node, and \textbf{Pallet\_SortOnly} is that only part-sorting operations are connected to the buffer node, and the buffer message is broadcast uniformly to every part-sorting node without differentiation. This refinement enables the model to focus on critical decisions and eliminate noise from non-relevant operations, and achieve better performance than the above two configurations. The best performance is achieved with \textbf{Pallet\_SortOnly\_Weighted}, in which the buffer message is discriminally broadcast to part-sorting nodes, and the edge weights are proportional to the different estimated pallet changes of different part-sorting nodes. We have also compared two edge weight strategies: \textit{cost-avoiding} and \textit{benefit-seeking}, the latter is represented by \textbf{Pallet\_SortOnly\_InverseWeight}. The \textit{benefit-seeking} strategy performs similarly to the baseline and worse than \textit{cost-avoiding}. We analyze that \textit{benefit-seeking} may lead to locally optimal but short-sighted choices while \textit{cost-avoiding} better mitigates high-cost decisions with a long-term consideration, resulting in more efficient scheduling.
% \begin{figure}[t]	
% 	\centering
% 	\includegraphics[scale=0.12]{AnonymousSubmission/LaTeX/Figure/Image/dd715e89a7e1cebdd95ce438f7ad55a.png}
% 	\caption{Simulation of the production line.}
% 	\label{fig_alg_compare}	
% \end{figure}
\section{Acknowledgments}
This work was supported in part by the NSFC (62325211, 62132021), the Fundamental Research Funds for the Central Universities (2042025kf0014), the Major Program of Xiangjiang Laboratory (23XJ01009), Key R\&D Program of Wuhan (2024060702030143).
\section{Conclusion}

This work addresses the Flexible Job-Shop Scheduling Problem with Limited Buffers and Material Kitting (FJSP-LB-MK) in manufacturing. We propose a DRL-based scheduling method using a heterogeneous graph neural network, which effectively models dependencies between machines, operations, and buffers. The method is validated on the synthetic dataset and real production line datasets, and compared with heuristics, constraint programming, and DRL methods. Experimental results demonstrate that our approach achieves a superior balance between performance and computational time, and ablation studies demonstrate the effectiveness of the key components in our method.

% In future work, we aim to extend our method to more complex, real-world production scenarios where buffer constraints vary significantly. We will explore applying our approach to practical production line tasks with diverse and dynamic buffer limitations, further optimizing scheduling strategies to meet a broader range of industrial challenges.

\bibliographystyle{IEEEtran}
\bibliography{IEEEabrv,example_paper}

\end{document}